# Uncertainty quantification for probabilistic machine learning in earth observation using conformal prediction.

[1] Geethen Singh [*], [2,3] Glenn Moncrieff, [4] Zander Venter, [5] Kerry Cawse-Nicholson, [6,7] Jasper Slingsby and [1] Tamara B Robinson

[1] Centre for Invasion Biology, Department of Botany and Zoology, Stellenbosch University, South Africa

[2] Global Science, The Nature Conservancy, Cape Town, 7945, South Africa

[3] Centre for Statistics in Ecology, Environment and Conservation, Department of Statistical Sciences, University of Cape Town, Private Bag X3, Rondebosch 7701, South Africa

[4] Norwegian Institute for Nature Research—NINA, Sognsveien 68, 0855 Oslo, Norway

[5] Carbon Cycles and Ecosystems, Jet Propulsion Laboratory, California Institute of Technology, Pasadena, CA, United States

[6] Department of Biological Sciences and Centre for Statistics in Ecology, Environment and Conservation, University of Cape Town, Private Bag X3, Rondebosch 7701, South Africa

[7] Fynbos Node, South African Environmental Observation Network, Centre for Biodiversity Conservation, Cape Town, South Africa

Corresponding Author [*]: Geethen.singh@gmail.com

**Highlights**

• Machine learning for Earth Observation (EO) data enhances decision support systems.

• Data uncertainty is crucial for decision-making. Popular methods lack reliability.

• Conformal prediction shows potential for enhanced Uncertainty Quantification (UQ).

• We review UQ in EO and demonstrate the use of introduced Earth Engine tools for UQ.

• The prospect for conformal prediction to advance EO is discussed.

**Abstract**

Unreliable predictions can occur when using artificial intelligence (AI) systems with negative consequences for downstream applications, particularly when employed for decision-making. Conformal prediction provides a model-agnostic framework for uncertainty quantification that can be applied to any dataset, irrespective of its distribution, post hoc. In contrast to other pixel-level uncertainty quantification methods, conformal prediction operates without requiring access to the underlying model and training dataset, concurrently offering statistically valid and informative prediction regions, all while maintaining computational efficiency. In response to the increased need to report uncertainty alongside point predictions, we bring attention to the promise of conformal prediction within the domain of Earth Observation (EO) applications. To accomplish this, we assess the current state of uncertainty quantification in the EO domain and found that only 20% of the reviewed Google Earth Engine (GEE) datasets incorporated a degree of uncertainty information, with unreliable methods prevalent. Next, we introduce modules that seamlessly integrate into existing GEE predictive modelling workflows and demonstrate the application of these tools for datasets spanning local to global scales, including the Dynamic World and Global Ecosystem Dynamics Investigation (GEDI) datasets. These case studies encompass regression and classification tasks, featuring both traditional and deep learning-based workflows. Subsequently, we discuss the opportunities arising from the use of conformal prediction in EO. We anticipate that the increased availability of easy-to-use implementations of conformal predictors, such as those provided here, will drive wider adoption of rigorous uncertainty quantification in EO, thereby enhancing the reliability of uses such as operational monitoring and decision making.

1. Introduction

The use of machine learning and Artificial Intelligence (AI) methodologies for geospatial applications (GeoAI) that employ Earth Observation (EO) data is crucial for monitoring advancements towards Sustainable Development Goals (SDGs) and other global accords, including the United Nations (UN) Convention on Biological Diversity (CBD) (Ferreira et al., 2020; Holloway et al., 2018; Pereira et al., 2013; Petrou et al., 2015). Notably, the Group on Earth Observations has identified which SDGs are quantifiable, to some extent, through EO data (Ferreira et al., 2020; Kavvada et al., 2020). Additionally, efforts have been undertaken to define more directly measurable variables that enhance biodiversity indicators (Skidmore et al., 2021), exemplified by the Essential Biodiversity Variables (EBV) (Pereira et al., 2013). The foundation of sustainable development hinges upon data-driven decision making that is partly informed by uncertainty information. If the uncertainty of the data underpinning a decision is too high, its utility for decision-makers diminishes. Similarly, if the uncertainty is not correctly quantified and presented, it may result in suboptimal decision outcomes.

## 1.1. Geospatial Artificial Intelligence (GeoAI) in ecology and environmental management

Historically, ecological observations have been made by experts in the field, over limited spatial extents with few or no revisits. The emerging field of geospatial artificial intelligence (GeoAI) holds great promise for revolutionizing conservation and environmental management (Janowicz et al., 2020; Song et al., 2023). It has the potential to enhance the capabilities of ecologists by increasing their field of view and frequency of observation at a lower cost than equivalent field work (Ball et al., 2017; Janowicz et al., 2020; Larson et al., 2020; Song et al., 2023). However, there exists a large gap between the research priorities of scientists and the practical requirements for informed decision-making (Müllerová et al., 2023). Among the factors that underscore this are end-user trust and the point-prediction nature of EO-derived datasets that have limited flexibility to meet the needs of end-users. For instance, there are regions of the world that are critically under-sampled, which may not be well characterized by models trained on data from the global North, reducing the suitability of the derived data and end-user trust (Ludwig et al., 2023).

Effectively addressing these challenges requires the recognised practice of proactive engagement with stakeholders through, for example, the clear communication of the limitations of the data being used (Müllerová et al., 2023; Paasche et al., 2022). To further narrow this knowledge-doing gap, a valuable addition to contemporary GeoAI systems involves using Uncertainty Quantification (UQ) techniques (Duncanson et al., 2022; Lang et al., 2023; Valle et al., 2023). The widespread application of AI systems will inevitably expose these systems to data beyond the scope of their training data, compromising system performance (Quinonero-Candela et al., 2008; Sugiyama and Kawanabe, 2012). For example, this can occur in low-frequency scenarios (Quinonero-Candela et al., 2008), such as the introduction of new alien plant species or climatic change resulting in an altered environmental template. In addition, differences in data sources and spatio-temporal context when making predictions compared to the model's training data can further challenge system performance during inference (Quinonero-Candela et al., 2008; Sugiyama and Kawanabe, 2012).

In these scenarios, UQ stands to benefit both data creators and data users. Discerning and flagging unreliable predictions can lead to a better understanding of a model's biases and errors leading to targeted data collection, the development of improved models and GeoAI systems with reduced uncertainty, wider adoption and improved efficiency. For data users, the communication of prediction uncertainty could enhance trust (Duncanson et al., 2022; Jacovi et al., 2021; Nicora et al., 2022), and help mitigate the negative consequences associated with the over-reliance on unreliable predictions that could lead to erroneous decision making (Lang et al., 2023; Zhao et al., 2023).

UQ methodologies, in certain instances, advocate for producing continuous probabilistic predictions, thereby affording increased flexibility in downstream applications. For instance, end-users can select a threshold probability better aligned with their spatio-temporal context and the associated real-world costs of omission and commission errors. Despite the benefits to be gained from UQ, a widely adopted framework that is reliable, flexible and easy-to-use is absent in the field of machine learning, let alone in EO (Duncanson et al., 2022; Paasche et al., 2022; Valle et al., 2023).

## 1.2. Uncertainty and its quantification in Earth Observation (EO)

Data acquired by satellites, including reflectance spectra, backscatter or waveform data contain inherent uncertainty owing to measurement noise, randomness, unpredictability in a system, sensor anomalies (for example Landsat-8 thermal calibration issues, Barsi et al., 2020), imperfect pre-processing steps (for example, atmospheric correction, orthorectification and terrain corrections) and partial data acquisition (For example, due to the scan-line error in Landsat-7 or the acquisition footprint of GEDI Paasche et al., 2022; Wang et al., 2021). These sources of uncertainty represent irreducible error and are denoted as aleatoric uncertainty (Gruber et al., 2023). In addition, uncertainties that arise from the lack of knowledge or understanding of a system, the selected modelling framework and through the stochastic nature of model fitting are cumulatively referred to as epistemic uncertainty (Gruber et al., 2023). Both categories of uncertainty are considered in this study.

Uncertainty is distinct from error/accuracy, uncertainty quantification defines the estimated distribution within which the true value lies and corresponds to the confidence in a prediction. Conversely, error is defined as the difference between an observed true value and a model prediction. The disclosure of uncertainty information alongside prediction error constitutes a complementary practice (Cohen, 1996). This synergy is underpinned by the inherent sparsity of error data, as it is quantified solely in the presence of reference data. In contrast, uncertainty information can be systematically reported for all pixels, regardless of the sparse availability of reference data. Moreover, in the context of classification or regression, accuracy and error metrics convey information for the overall quality of a dataset, whereas some uncertainty methods can provide pixel-level information on the prediction quality.

While conventional ensemble methods that rely on assessing the (dis)agreement among multiple model predictions are widely employed in predictive modelling for pixel-wise UQ, they only capture epistemic uncertainty (Valle et al., 2023). Moreover, their operational deployment is associated with high computational costs given the need to train multiple models. Analogous to ensemble methods employed in classification and regression tasks, the derivation of pixel-wise uncertainty in regression tasks can be achieved through quantile regression. However, it does not expressly guarantee the coverage stipulated by the quantile levels. Specifically, a quantile regression model incorporating predictions at the lower 5th and upper 95th quantiles often proves inadequate in encompassing the true value within the prediction interval for a majority (90%) of instances (Romano et al., 2019). Design-based area estimation and bootstrapping represent prevalent methodologies for establishing confidence intervals for accuracy and area estimates (Olofsson et al., 2014). Nevertheless, both methodologies, by design, fall short by not providing pixel-wise uncertainty information and are not yet readily accessible owing to the specialised knowledge requirement and lack of tooling and software.

### 1.3. Conformal prediction: A potential solution

Conformal prediction is a mathematical framework that can quantify both epistemic and aleatoric uncertainty in combination (Christoph Molnar, 2023). It is amenable to integration with any prediction model and dataset, irrespective of its statistical distribution, while adhering to a pre-specified confidence level (Angelopoulos and Bates, 2023; Christoph Molnar, 2023; Vovk et al., 2005). This translates to the provision of uncertainty estimates with a constrained error rate or tolerance level. For example, if a 95% confidence level is specified, the conformal predictor will provide a prediction region that contains the true value with a 95% probability (for classification problems, this prediction region corresponds to a set of labels, a multilabel prediction) (Angelopoulos and Bates, 2023; Christoph Molnar, 2023). This coverage guarantee is referred to as the validity property of conformal predictors and remains conspicuously absent from all other pixel-wise UQ methodologies, except under certain distribution assumptions (Manokhin, 2022a; Shafer and Vovk, 2008; Vovk et al., 2005). The fulfilment of the validity property of conformal prediction hinges on the exchangeability assumption being satisfied. Exchangeability signifies that the data utilized for calibrating the conformal predictor could be swapped with the test data without affecting the probability distribution of the target variable to be estimated.

Fulfilling the validity property in isolation proves inadequate because a broad prediction region or a set encompassing all candidate classes will meet the coverage criteria but will be uninformative. Hence, the statistical efficiency pertaining to the length or width of prediction regions in classification and regression tasks becomes important, necessitating smaller prediction set sizes and narrow prediction intervals. In instances where a prediction fails to reach the stipulated confidence level, a null prediction may be provided. Alternatively, when the prediction region is too large to be informative, the corresponding prediction can be flagged for human intervention (Christoph Molnar, 2023). In this way, conformal prediction can serve as a quality control system devoted to dependable predictions.

Conformal prediction has been attracting growing attention, however, it has yet to be widely adopted within the domain of GeoAI (Norinder and Lowry, 2023; Valle et al., 2023). A few studies have ventured into the application of conformal prediction to EO data, with a principal focus on verifying its validity and efficiency property in the context of EO data for classification tasks across small extents (Norinder and Lowry, 2023; Valle et al., 2023). This work extends previous work by i) providing easy-to-use tools that can be easily integrated with current GEE workflows, and ii) demonstrating the scalability, and computational efficiency of the introduced conformal regression and classification tools at local to global extents. By doing so, we aim to promote the widespread adoption of conformal prediction to quantify uncertainty for probabilistic machine learning in EO. This aim is pursued by i) Systematically assessing the current status of UQ in EO by commencing with a review of the UQ methods employed for datasets catalogued in the Google Earth Engine (GEE) and GEE community catalogues (Roy et al., 2023). ii) Demonstrating practical use through three distinct case studies, each covering an increasing geographic extent, from local to global, and includes both regression and classification tasks for GEE feature (vector data) and image collections (raster data) applicable to both traditional and deep learning-based machine learning workflows. iii) Next, we discuss the opportunities offered by conformal prediction techniques that are likely to advance EO and future operational systems. iv) Finally, to facilitate the use of conformal prediction techniques within the EO community, this research contributes to the availability of open-source tools and tutorials. Specifically, Python and JavaScript modules are made accessible, intended to seamlessly integrate with pre-existing GEE workflows.

## 2. Methods

**2.1. Assessing the status of Uncertainty Quantification (UQ) in Earth Observation (EO).**

To assess the current trends and status of different UQ methods in EO, we examined all machine learning derived datasets in the GEE and the GEE community catalogue with the latest update on the 2 November 2023 considered (Roy et al., 2023) (Table 1, Table A1, Table A2). These data catalogues were selected since they contain commonly used datasets with a national to global coverage. For the core GEE catalogue, "machine learning", "uncertainty" and "UQ" keywords were used to find and filter all machine learning derived datasets that were reviewed. If a dataset was not tagged with one of three keywords or quantified uncertainty but was derived using expert rule-based systems or statistical models, they were not returned in search results and were therefore not considered. In addition to examining the band information, the full-text research paper was examined to determine if i) machine learning was used to derive the dataset, ii) uncertainty was quantified for the derived dataset and, if it was quantified, iii) which method was employed. Overall, 241 datasets were assessed.

**2.2. Demonstrating the utility of conformal predictors.**

Three case studies were selected to demonstrate the broad utility of conformal prediction. Specifically, the case studies look at quantifying uncertainty for invasive tree species mapping for a region in South Africa, canopy height estimation based on Global Ecosystem Dynamics

Investigation (GEDI) for Africa (Dubayah et al., 2020), and land cover classification using the global Google Dynamic World dataset (Brown et al., 2022). We consider applications with small (<1500 instances) to large datasets (>110M instances), for both classification and regression and both GEE image and feature collections, which are used in deep-learning models and traditional machine learning models, respectively. The steps taken to produce the uncertainty information in each case study is summarised below (Figure 1). For both classification tasks (2.2.2. and 2.2.4.), we use the least ambiguous set-valued conformal classifier method (Sadinle et al., 2019). For the canopy height regression case study (2.2.3.), we use conformal quantile regression (Romano et al., 2019). Both methods are described in detail below (2.2.1)

**2.2.1. Conformal prediction: The six steps**

Practically, producing uncertainty estimates using conformal prediction involves six steps: the initial procedural phase entails the partitioning of a given reference dataset into three subsets, namely the training set, the calibration set, and the test set (Angelopoulos and Bates, 2023; Christoph Molnar, 2023). Subsequently, a predictive model is trained on the training set, after which it is deployed to estimate the class probabilities or regressed values within the calibration and test sets. In the third step, during the calibration phase, each calibration instance is scored based on its nonconformity with the true value. An example of a simple but common scoring function for classification tasks is hinge loss and encompasses the subtraction of one from the classifier-produced probability-like scores (Angelopoulos and Bates, 2023; Lei et al., 2018). Next, as part of the calibration stage, these nonconformity scores are used to compute a probability threshold corresponding to the user-defined confidence level (1-alpha) after a finite-sample correction (Equation 1, Angelopoulos & Bates, 2023; Sadinle et al., 2019).

$$qLevel = \frac{ceil\left((1 + nCal) * (1 - \alpha)\right)}{nCal} \quad (1),$$

Where, *qLevel* corresponds to the adjusted quantile level, *alpha (α)* corresponds to the proportion of acceptable errors or tolerance level and *nCal* denotes the size of the calibration set. In the fifth step, post-calibration and during the inference stage, the computed probability threshold value is used in the creation of class sets for each test instance. All classes with nonconformity scores that are greater than or equal to the probability threshold value are included in the output prediction set (Shafer and Vovk, 2008). This corresponds to the inclusion of class labels in the prediction set if their associated confidence exceeds a desired and user-specified confidence level (Christoph Molnar, 2023).

In the final step, the test set is used in the evaluation stage to assess the validity and efficiency of the calibrated conformal predictor by computing the empirical marginal coverage and the average set size, respectively. For instance, if the specified confidence level is 95%, then ~95% of the prediction sets for the out-of-sample test instances, across all classes, should include the true class. If the coverage deviates from the specified confidence level, this implies a violation of the exchangeability assumption through, for example, quantifying uncertainty in a region that was not represented in the training and calibration data. This assumption assumes the nonconformity scores between the calibration and test sets are permutation invariant and thus, their ranks being uniformly distributed (Angelopoulos and Bates, 2023; Shafer and Vovk, 2008).

For regression tasks, the six steps remain unchanged with the exception of the scoring function used to generate the nonconformity scores and the method used to evaluate the prediction regions. The most used scoring function in regression computes the absolute residual for each calibration and test instance. During the inference stage, the absolute residual corresponding to the user-specified confidence level is added and subtracted from the mean prediction values to provide an upper and lower bound, respectively. A drawback of this simple scoring function is the lack of adaptability i.e., all prediction intervals have the same width. Therefore, conformal quantile regression has been introduced to provide adaptability whereby more difficult prediction instances have wider intervals than easier prediction instances (Angelopoulos and Bates, 2023; Romano et al., 2019).

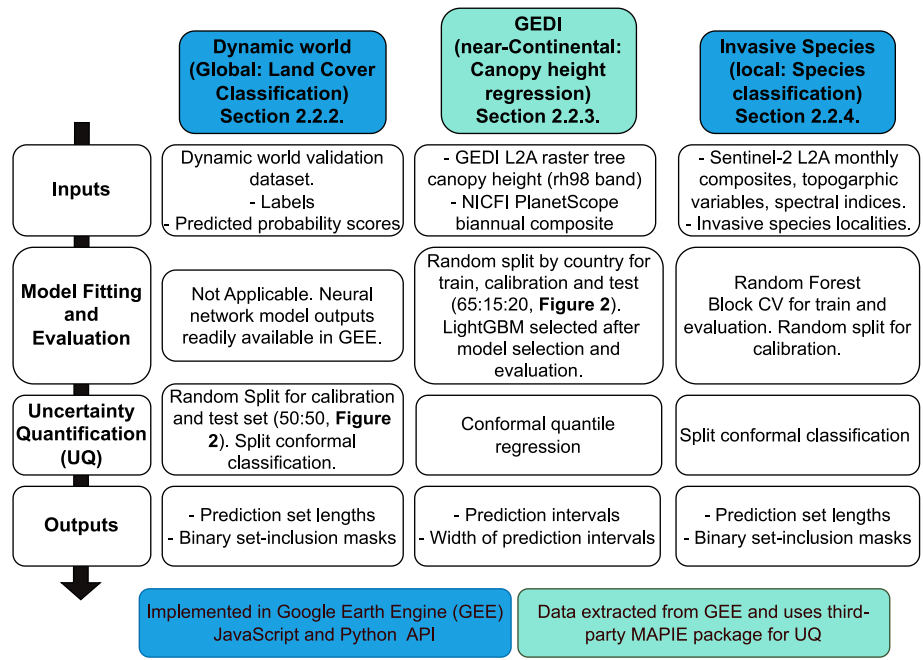

**Figure 1.** Summarising the workflow for each of the three case studies including uncertainty quantification using the Google Earth Engine (GEE) JavaScript code editor API (blue) and the GEE Python API with the Python MAPIE package (turquoise).

**2.2.2. Global Google Dynamic World (classification)**

Google's Dynamic World dataset is a near-real time land cover product produced for every Sentinel-2 scene with less than or equal to 35% cloud cover. As part of the neural network based landcover model output, a per-class probability band is produced in addition to a discrete class output band corresponding to the class with the maximum probability. The discrete labels for 2020 were assessed to have an overall agreement of 73.8 % against expert-derived reference labels (Brown et al., 2022). To quantify uncertainty, the released globally distributed dynamic world validation set comprised of 409 reference label images (512 x 512 pixels) and corresponding predicted per-class probability bands (refer to Brown et al., 2022, for details on the sampling design) were randomly split into 80% calibration and 20% test data (Figure 2A). The calibration set was used to calibrate the least ambiguous set-valued conformal classifier (2.2.1), while the test set was used to evaluate the empirical marginal coverage and average set size. Assuming the calibration set scores are exchangeable with the scores derived from any output dynamic world scene, the calibrated conformal classifier can be applied to produce prediction sets that will include the actual class with a high (0.9) probability. Probability threshold values are computed for various confidence levels (0.7 ≤ (1-$α$) ≤ 0.95, in 5% intervals, Table A3).

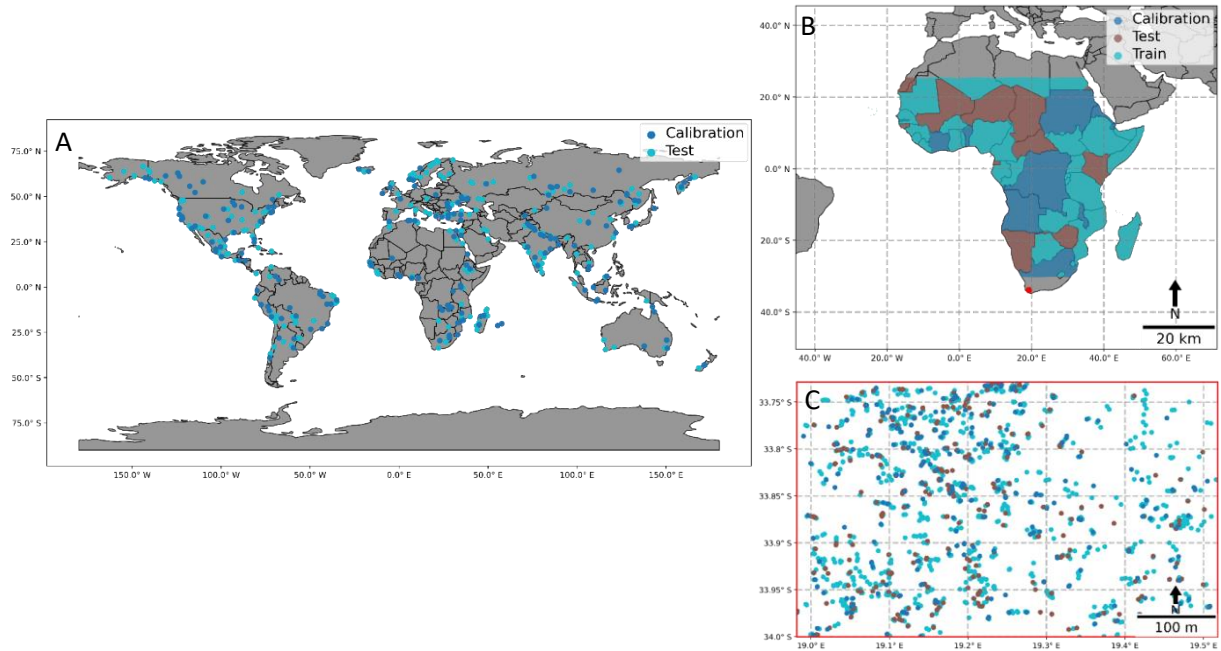

**Figure 2.** The distribution of train (if applicable), test and calibration samples for a) Land cover classification using Dynamic World, randomly split (80:20) calibration and test samples derived from the 409 out-of-sample validation samples made available by Google. Each sample represents an image with 512 x 512 pixels. b) Canopy height estimation using GEDI randomly split by country for the train (>50M points), test (>31M points) and calibration (>32M points) samples (65:20:15). The countries considered have been limited to the extent of the NICFI Africa PlanetScope dataset and c) Invasive tree species classification randomly split (65:20:15).

### 2.2.3. Continental GEDI canopy height (regression)

NASA's Global Ecosystem Dynamics Investigation (GEDI) is a space-based laser altimeter with a full-waveform detector that captures the vertical structure and distribution of vegetation, a proxy for biomass and tree canopy height (Duncanson et al., 2022). These volumetric tree-stand variables are captured in 100 relative height (rh) bands. For example, the selected rh98 response band corresponds to the height at which 98% of energy is returned to the detector from a 25x25 m footprint area (Dubayah et al., 2020). In this case study, the canopy height product made available in GEE was combined with the former 2020 biannual Visible and Near-Infrared (VNIR) predictors from the NICFI PlanetScope data (Team, 2017). Predictors were extracted from GEE for each GEDI instance overlapping the Africa PlanetScope extent. Next, each country in the extent was randomly assigned to a train, calibration, or test set (Figure 2B). The train set was used to fit a decision tree based Light Gradient Boosting Machine (LightGBM) quantile regression model, thereafter, the calibration set was used to calibrate the quantile regressor. This enabled the provision of canopy height point estimates, together with a prediction interval that contains the actual canopy height value, as measured by GEDI, with a high probability (95%).

**2.2.4. Local Invasive tree species discrimination (classification)**

This case study presents a novel analysis and diverges from the previous Dynamic World case study based on data composition, the extent of interest, and the modelling workflow. Here, the dataset consists of georeferenced locations of species within a GEE feature collection instead of dense image labels like Dynamic World. We also demonstrate an entire workflow that involves model fitting, unlike Dynamic World which only involves UQ owing to the readily available class-wise probability outputs on GEE.

A combination of high-resolution Google Earth Imagery and familiarity with the distribution of the dominant invasive alien tree species in the local area of interest (Figure 2C, The Western Cape, South Africa) (Meijninger and Jarmain, 2014), was used to sample the invasive tree localities and its surrounding landcover. The invasive tree species include Acacia, Eucalyptus and Pinus species, commonly known as wattles, gums and pines respectively. The region is located in the water scarce Western Cape province of South Africa that receives ~380 mm rainfall per year (Holden et al., 2021). One of the major threats to water security in the upper Berg and Breede catchments is invasive alien trees (Holden et al., 2021; Meijninger and Jarmain, 2014). This case study represents a common scenario whereby managers are interested in invasive tree species within their jurisdiction of influence/mandate. Moreover, the limited size of the dataset (<1500 instances) and small extent of interest corresponds to a common set of conditions under which satellite remote sensing is used. Similar pilot studies are an important preliminary to scaling up the application of satellite based mapping and an important milestone to develop reliable large-scale operational monitoring programmes.

The mapping workflow used in this case study (Figure 1), included, feature preparation, model evaluation and uncertainty quantification. During feature preparation, Sentinel-2 level 2A median monthly composites were created after performing cloud and cloud-shadow masking based on s2cloudless (Skakun et al., 2022), gap filling and stacking additional spectral indices and topographic covariates that included, elevation, slope, aspect, topographic position index, continuous heat insolation index and derived location variables (i.e., a rotated coordinate variable that reduces the effects of spatial autocorrelation whilst accounting for spatial patterns Møller et al., 2020). Next, during the model training and hyperparameter tuning phase, a random forest model was trained using a 10-fold spatial cross validation approach. Here, the coordinates of the input tree species localities were clustered and split into folds to limit the effects of spatial autocorrelation. A summed confusion matrix and the cross-validation accuracy statistics are returned. Finally, for the calibration of the conformal predictor and the quantification of uncertainty, we used the least ambiguous set-valued conformal classifier, discussed above (2.2.1), with a model trained on the entire train set. The calibrated classifier can then be used to obtain pixel-wise sets and the corresponding length of the prediction sets with a 90% confidence level.

### 3. Results

**3.1. The status of UQ in Earth Observation (EO)**

Uncertainty is seldomly considered in EO with only 17 of the 87 (20%) reviewed datasets in the GEE catalogues citing studies that quantified uncertainty (Figure 3). This observation underscores the importance of UQ frameworks that are easy to use and support a wide range of machine learning tasks. For the 17 studies, quantile regression is commonly used for regression tasks while model ensembles that capture variance in probability-like scores are commonly used for classification tasks. Bootstrapping and design-based area estimates are commonly used to provide confidence intervals around accuracy scores and area coverage, respectively. Notably, many datasets that employ UQ relate to the estimation of forest structure and carbon sequestration (Table 1).

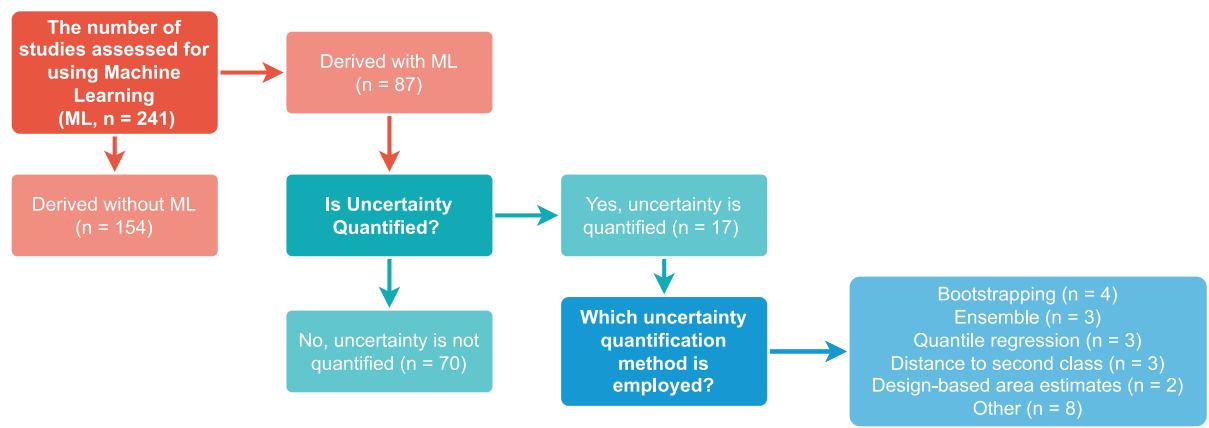

**Figure 3.** A schematic of the classification of each assessed dataset (n = 241) from both the Google Earth Engine (GEE) catalogue and the GEE community catalogue. Only the datasets that used machine learning were considered from the main GEE catalogue. For the community catalogue, all datasets comprising the catalogue up to the 2 November 2023 update were considered. Five studies used more than one UQ method.

**Table 1.** An overview of the uncertainty methods used in the 17-machine learning-derived datasets made available through the Google Earth Engine (GEE) data catalogue and the GEE community catalogue. The datasets shown here represent 20% of the total machine learning-derived datasets examined.

| Dataset | Uncertainty Quantification method | Spatio-temporal coverage | Reference |
| --- | --- | --- | --- |
| Soil carbon storage in terrestrial ecosystems of Canada | Quantile regression | National-Canada (N/A) | (Sothe et al., 2022) |

| Irrecoverable carbon in Earth's ecosystems | Standard error of the uncertainty layers of the used datasets. | Global (2010, and 2018) | (Noon et al., 2022) |
|---|---|---|---|
| Soil Grids 250m v2.0 | Quantile regression | Global (N/A) | (Poggio et al., 2021) |
| Global Mangrove Project | Bootstrapping for confidence intervals around accuracy statistics | Global (1996, 2007-2010, 2015- 2020) | (Bunting et al., 2022) |
| Land Change Monitoring, Assessment, and Projection (LCMAP) v1.3 | Model Quality Flags includes persistent snow, insufficient data and clear conditions and Sample based area estimates. | National-CONUS (1985-2021) | (Brown et al., 2020) |
| ETH Global Sentinel-2 10m Canopy Height (2020) | Negative log likelihood loss function for aleatoric uncertainty and ensemble predictions for epistemic uncertainty. | Global (2020) | (Lang et al., 2023) |
| High Resolution Tree Species Information for Canada | Distance to second class (DS2C) based on 100*(1-nVotesC2/nVotesC1) | National-Canada (2019) | ((Hermosilla et al., 2022) |
| Canada Landsat Derived Forest harvest disturbance 1985-2020 | DS2C based on 100*(1-nVotesC2/nVotesC1) | National-Canada (1985-2020) | (Hermosilla et al., 2016) |
| Rangeland Analysis Platform layers (rangeland fractional cover) | Ensemble based prediction variance | National-CONUS (2019) | (Allred et al., 2021; Jones et al., 2021; Robinson et al., 2019) |
| Ensemble Source Africa Cropland Mask 2016 | Sample based area estimates | Continental-Africa (2016) | (Nabil et al., 2022) |
| Highly Scalable Temporal Adaptive Reflectance Fusion Model (HISTARFM) database | Kalman filter | National-CONUS (2009-2021) | (Moreno-Martínez et al., 2020) |
| Global Photovoltaics Inventory (2016-2018) | Custom mechanistic approach which makes distribution assumptions and bootstrapping. | Global (2016-2018) | ((Kruitwagen et al., 2021) |
| Canada Landsat Derived Wildfire disturbance & Magnitude 1985-2020 | (DS2C) based on 100*(1-nVotesC2/nVotesC1) | National-Canada (1985-2020) | (Hermosilla et al., 2016) |
| RADD Forest Disturbance Alert | Probabilistic mapping using Gaussian mixture models and Bayesian methods | Global (2019-2020) | (Reiche et al., 2021) |
| iSDASoil | Quantile regression and bootstrapping. | Continental – Africa (2021) | (Hengl et al., 2021) |
| Global urban projections under SSPs (2020-2100) | Ensemble based prediction variance | Global (2020-2100) | (Chen et al., 2020; Gao and O'Neill, 2020) |
| Murray Global Intertidal Change | Quality flags that contain the number of input pixels for modelling. | Global (1984-2016 in 3-year intervals) | (Murray et al., 2019) |

## 3.2. Applying conformal prediction

The overall lack of UQ in the datasets considered supports our case for conformal prediction as an easy-to-use and robust UQ framework that is suitable for machine learning derived datasets in EO. We demonstrated its use for small to large dataset sizes encompassing local to global extents and both regression and classification tasks.

**3.2.1. Global Google Dynamic World (classification)**

For classification tasks, we represent pixel-wise uncertainty as the number of classes included in a pixels' prediction set. Highly uncertain predictions can either be represented with an empty set (length equal to zero) or a large multi-label set (length closer to the total number of candidate classes; for instance, nine for Dynamic World). A multi-label set suggests that the prediction model is finding it challenging to distinguish between several possible class labels at the desired confidence level. Higher desired confidence in the prediction leads to larger prediction sets or intervals (analogously to how higher desired confidence in parameter estimates lead to larger confidence intervals). Although such a prediction is not incorrect per se, it is inconclusive, and human intervention would be required to derive the true label. Empty set predictions are examples where the model could not assign any label, typically meaning that the example was very different from the data the model was trained on. Conversely, the most confident predictions are shown with set lengths of one. For instance, inland water and built-up predictions are among the most reliably mapped land cover classes (Figure 4A-B), whereas object boundaries typical of transition and seasonal areas are associated with higher prediction uncertainty and larger prediction set lengths (Figure 4B).

3.2.2. **Continental GEDI canopy height (regression)**

For the canopy height regression task (test set RMSE = 3.30m), we represent uncertainty as the difference between the upper and lower prediction bound, referred to as prediction interval. The prediction interval contains the actual canopy height, as based on GEDI, with a 95% probability (empirical marginal coverage, 95.15% ± 0.07). Prediction intervals with a greater width are representative of high prediction uncertainty. The average prediction interval width is 9.28m ± 0.03m. Higher canopy height (Figure 5A) corresponds to wider prediction intervals and greater uncertainty (Figure 5B), but when one looks at water systems and pans there are instances that deviate from this generalisation (yellow regions in Figure 5C). There is a marked increase in uncertainties in the context of water systems, and pans such as the Sua salt pan in Botswana and the Namibian Etosha pan (Figure 5B, red boxes), and the Terene Desert shared between Niger and Chad (Figure 5C, white box). Moreover, diagonal image artefacts (Figure 5B, feint lines), comprising aleatoric uncertainty due to seamlines, in central Africa also show a similar deviation.

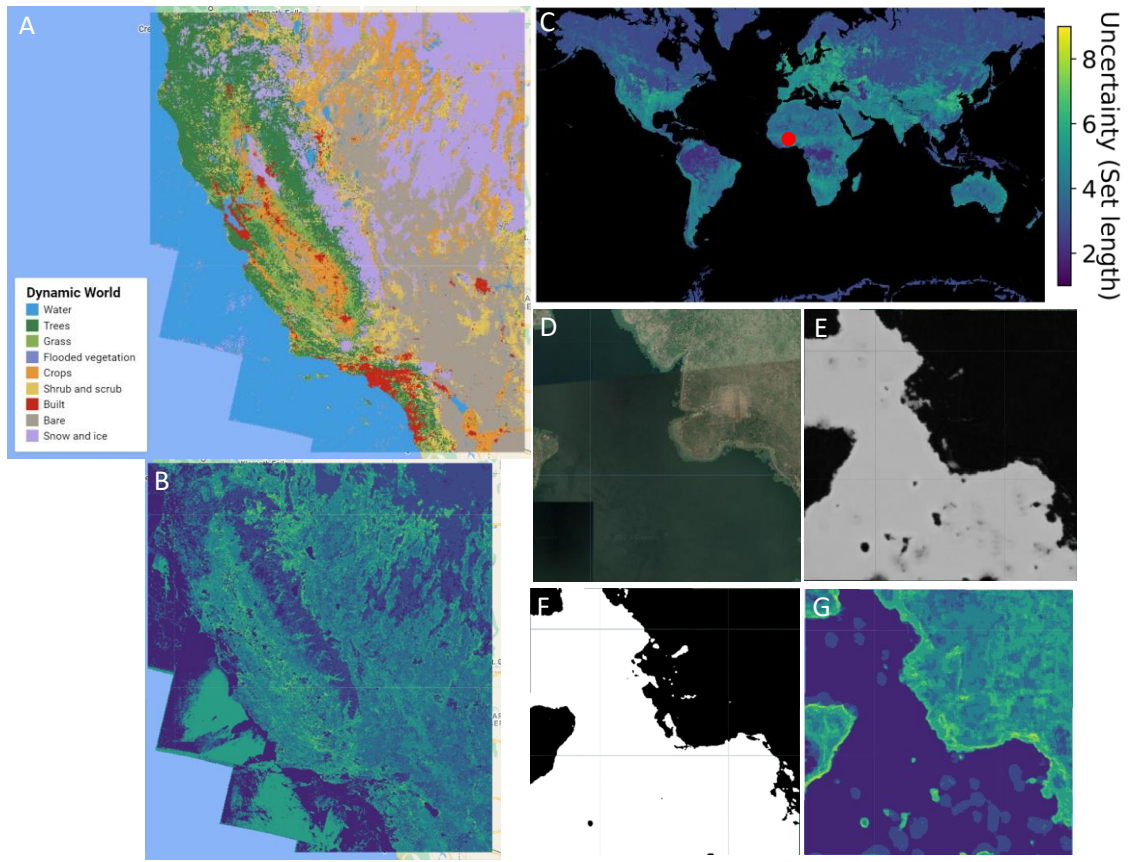

**Figure 4.** The a & b) Dynamic World land cover classification with associated uncertainty over California, United States of America showing confident inland water, built-up, snow and ice predictions. c) The global distribution of uncertainty for the first non-null land cover image in 2020. d-g) A high-resolution Google Earth Image (red point in C) with corresponding e) probability of water, f) the prediction sets that include the water class and g) the length of the prediction sets. Empty set predictions (length = 0) are not shown. Interactively explore the Dynamic World dataset with accompanying uncertainty information or quantify uncertainty for any Dynamic World scene here.

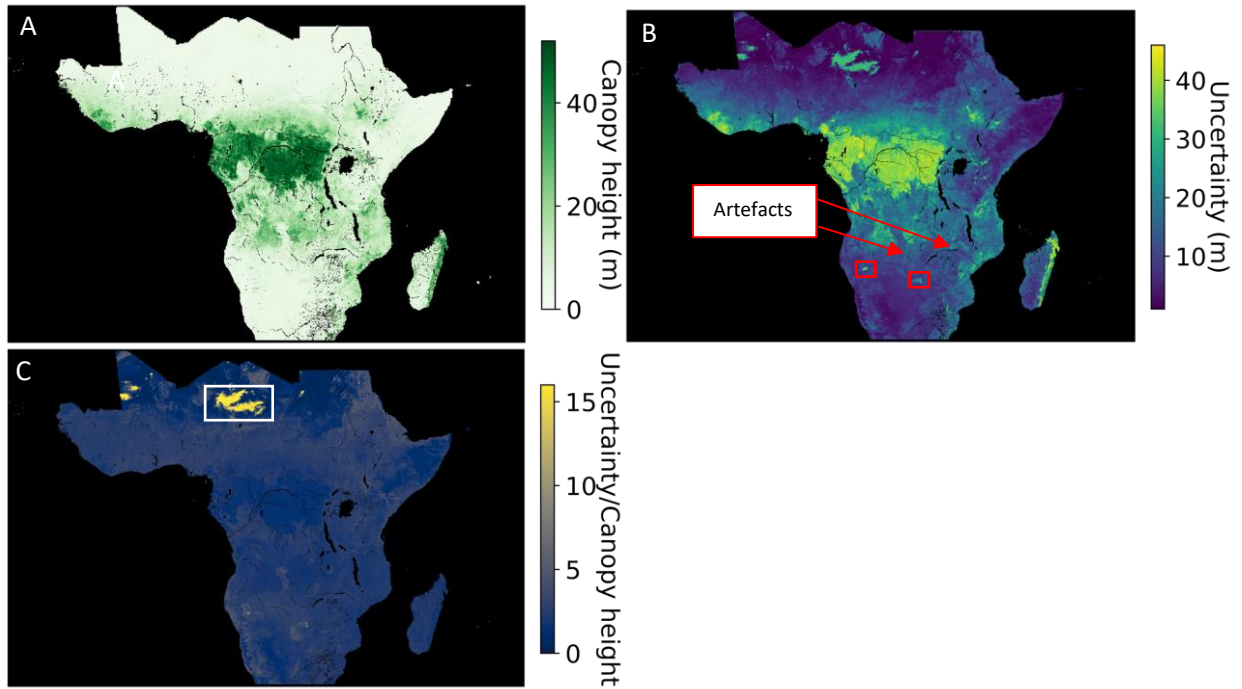

**Figure 5.** The distribution of a) tree canopy height as estimated from GEDI and NICFI PlanetScope data, b) associated prediction intervals with a 95% confidence level, highlighting diagonal linear artefacts and large prediction intervals for pans (red boxes) and c) the ratio between the two (a & b, uncertainty/Canopy height), highlighting an area in the Terene desert with high uncertainty and low canopy height. The data can be interactively explored [here](here).

**3.2.3. Local Invasive tree species discrimination (classification)**

The invasive tree species classifier has an average set size of 1.53 and an empirical marginal coverage of 0.92. The small set size suggests that the model produces mostly confident predictions and that the time series covariates are highly suited for discriminating and localising pine, wattle, and gum tree species. The uncertainty information (prediction sets) of, for example, pine trees could be used to prioritize the inspection sites that may contain pine (set length > 1) or to find pine infested areas, with a 90% probability, for intervention efforts (set length = 1, Figure 6).

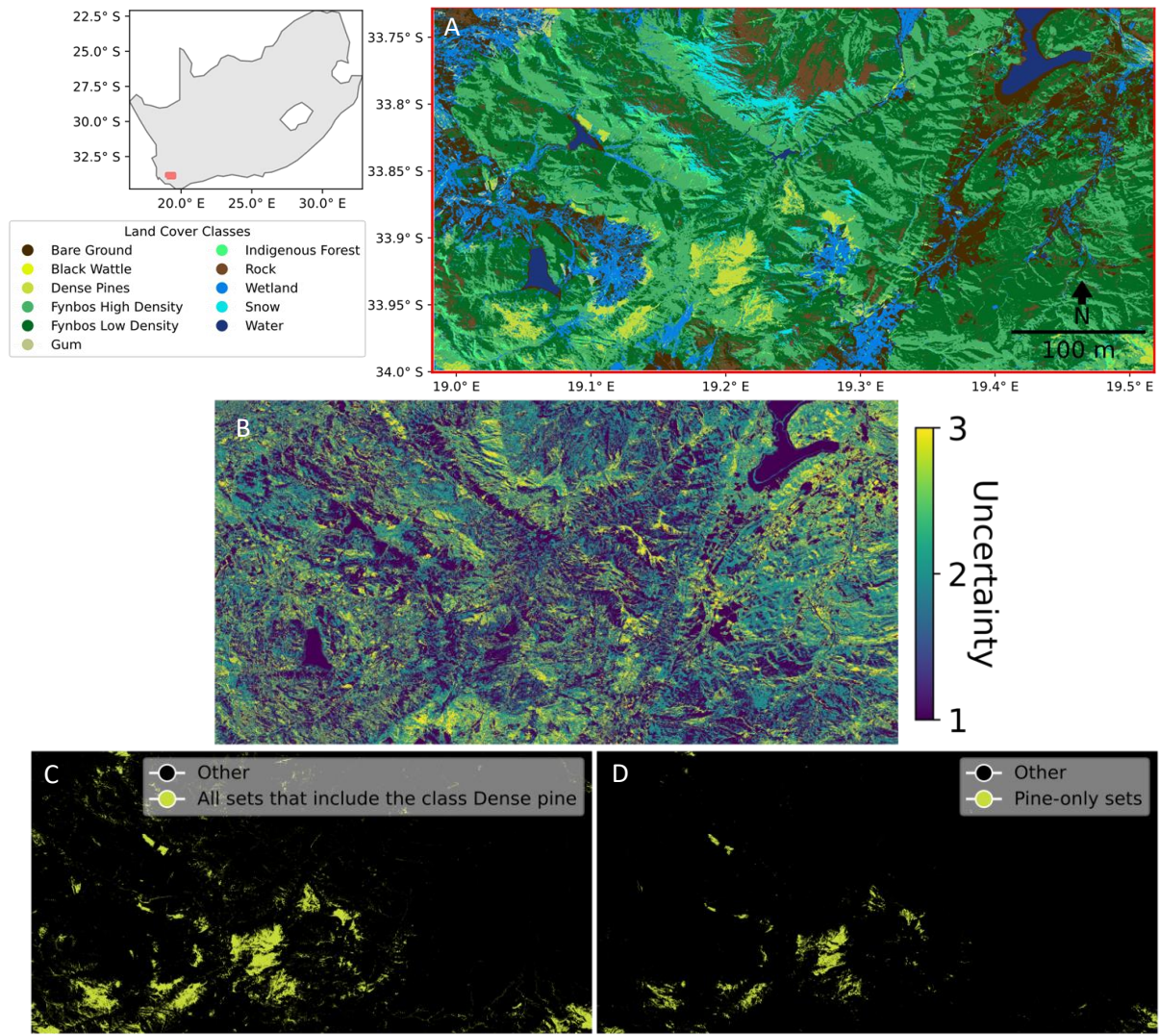

**Figure 6**. The a) discrete classification of invasive tree species and its surrounding land cover, b) associated uncertainty as presented by the pixels' prediction set length, c) all pixels that include pine species along with other classes within their predictions sets (set length > 1), and d) pixels that only include the Dense pine class in their prediction sets (set length = 1) with a 90% confidence level. Training, calibration and test data only covered natural lands and hence agricultural and urban areas are associated with high uncertainty.

## 4. Discussion

Uncertainty Quantification plays a crucial role in machine learning, serving two fundamental purposes. Firstly, it serves as an effective and consistent method for communicating the quality of predictions to end-users, and secondly, as a measurable parameter facilitating the comparison and integration of diverse estimations of a target variable. Studies that use machine learning to generate predictions based on EO data do not routinely report uncertainty for several reasons, including a lack of consensus in methodologies, lack of easy-to-use tools, and a lack of access to computational power required for some methods (Duncanson et al., 2022; Valle et al., 2023). We have demonstrated the utility of conformal prediction to quantify uncertainty in a robust and scalable manner relevant to EO-based studies that rely on classification and regression tasks.

## 5.1. The status of UQ in EO

Following the examination of the two data catalogues, we found that a minority (36%) of the derived datasets had been produced using machine learning techniques. This can be attributed to the early but growing prevalence of large-scale labelled EO datasets and expertise in applying machine learning methods to EO data. Machine learning derived datasets are becoming commonplace and embedded in decision making (Ferreira et al., 2020; Holloway et al., 2018; Kavvada et al., 2020; Pereira et al., 2013; Petrou et al., 2015). Thus, it is important that we adopt easy-to-use and robust UQ methods to ensure that the quality of the generated datasets is not ambiguous.

Of the datasets derived through machine learning, 70 of the 87 datasets (80%) did not quantify uncertainty. Notably, some research efforts conflated the concepts of error and uncertainty (for example, Sexton et al., 2013; Venter & Sydenham, 2021). It is therefore crucial to distinguish between prediction uncertainty and error. Prediction uncertainty pertains to the confidence that a prediction accurately reflects the quantity being measured. This is achieved by presenting a range of potential values associated with the prediction (Cohen, 1996). Whereas prediction error or accuracy reflects the disparity between actual and predicted values and partly contributes towards uncertainty. While error and uncertainty are distinct, they are both complementary to understanding model output data (Cohen, 1996). In 20% of the studies where UQ methods were utilized, it appears that the selection criteria often favoured ease of implementation. Moreover, an evaluation of the quality (validity and efficiency) of the quantified uncertainty was notably lacking, except for its partial consideration in a single study (Lang et al., 2023).

A substantial proportion of the studies that quantified uncertainty, were directed toward estimating sequestered carbon. This trend can be ascribed to the critical role of uncertainty quantification in carbon crediting. Various regulatory protocols (e.g., the Climate Action Reserve), have adopted a methodology wherein carbon credits are not allocated based on the median (point) estimate of sequestered carbon, but rather on the lower boundary of a prediction interval (Potash et al., 2023; Reserve, 2022). This approach is adopted to mitigate the risk of granting credits for carbon that has not genuinely been sequestered. While ensemble and quantile regression methods of quantifying uncertainty are generally accessible and straightforward to implement, it is important to note that neither of these approaches offer valid coverage guarantees under any data distribution and may result in a lower bound that is either over-conservative or overly optimistic.

## 5.2. A comparison of conformal prediction to other UQ methods used in EO

Conformal prediction exhibits advantageous attributes for diverse machine learning applications, characterized by its model-agnostic nature, distribution-free characteristics, inherent validity, and computational efficiency, alongside straightforward implementation protocols. These attributes confer distinct advantages relative to alternative methodologies employed for UQ in EO (refer to Table 1). Notably, conformal prediction enables the provision of valid quantitative uncertainty information, a capability surpassing that of quality flags that provide a qualitative assessment of the reliability of a prediction based on heuristics such as the number of pixels used to create a prediction. The Distance to Second Class (DS2C) method, employed for uncertainty quantification in random forest models, is exclusively applicable to decision-tree based models (Hermosilla et al., 2022, 2016). This heuristic notion of uncertainty, although limited to decision-tree based models, could concurrently serve as the foundation of a nonconformity score function that could be investigated in future work.

In contrast to UQ methodologies that requires the training of multiple models, such as those employed in quantile regression and ensemble strategies (Nicora et al., 2022; Shaker and Hüllermeier, 2020), the split-conformal paradigm obviates the need for retraining the predictive model. Techniques like bootstrapping and ensemble methodologies are computationally costly and reliant upon differences in replicate model predictions to reflect high uncertainty (Nicora et al., 2022). The use of split-conformal prediction in scenarios characterized by extensive labelled datasets, representative of Dynamic World, will be more efficient if probability thresholds are made publicly available for various confidence levels. This will circumvent the calibration phase for new images or new users (refer to Table A3). Furthermore, the continuous development of conformal prediction through active research, ensures its sustained pertinence and efficacy with support for a diverse and expanding array of tasks that include regression (Romano et al., 2019; Sesia & Candès), classification (Angelopoulos and Bates, 2023; Valle et al., 2023), time series analysis (Garza and Mergenthaler-Canseco, 2023; Stankeviciute et al., 2021), semantic segmentation (Teng et al., 2022; Wieslander et al., 2020), diffusion models in generative AI (Teneggi et al., 2023), as well as reinforcement learning and anomaly detection (Gupta and Kahou, 2023).

A limitation impeding the integration of conformal prediction into probabilistic machine learning for EO pertains to the lack of accommodation of post-processing procedures (for example, post classification filtering using erosion and dilation operations to remove isolated pixels or conditional radiance fields for semantic segmentation) that are often applied to the model's output without adjusting the underlying probability scores, resulting in a disparity between the improved model outputs and the unadjusted probability-derived non-conformity scores. Consequently, post-processing of predictions is likely to result in conservative prediction regions. To ensure statistical efficiency of the conformal predictor, it would therefore be advantageous if post-processing steps could be performed as pre-processing steps when possible or dropped. Another drawback arises when the nonconformity scores derived for the calibration data lack exchangeability with those obtained during inference, potentially leading to prediction intervals with coverage below the user-defined confidence level (1-alpha). For example, this may occur under data drift scenarios resulting from changing atmospheric conditions, land use land cover change or climate change. This necessitates recalibration. Nevertheless, recent research has explored extensions to conformal prediction under distribution shifts, encompassing covariate and label shifts (Gibbs and Candès, 2022; Gibbs and Candes, 2021; Tibshirani et al., 2019).

In summary, conformal prediction is a compelling methodology with considerable potential to enhance the precision and dependability of machine learning systems. Moreover, ongoing research is likely to resolve or mitigate certain drawbacks associated with this approach. For instance, Mondrian conformal prediction or class-conditioned conformal prediction was introduced to extend the coverage guarantee to encompass distinct classes or strata. This adjustment more closely aligns with the notion of conditional coverage, which mandates that each individual instance meets the prescribed coverage criteria (Löfström et al., 2015; Vovk, 2012). Split conformal prediction provides an assurance of coverage on a population level, indicating marginal coverage across the entire dataset. This implies that certain subpopulations, delineated by categories or strata, may surpass the user-defined coverage, while others may fall short of the stipulated criterion (Löfström et al., 2015; Vovk, 2012). Mondrian conformal prediction may be beneficial in cases whereby errors for specific classes are associated with higher real-world costs. For example, rare species or invasive species occurrences.

### 5.3. How could UQ advance applied Earth Observation?

Operational machine learning systems are characterised by their capacity to consistently deliver datasets derived from satellites in an ongoing manner, thereby facilitating decision making. The successful integration of operational GeoAI systems and decision support systems depends on overcoming challenges that address the disparity between existing information provided by EO-derived products and the needs outlined by envisioned data users such as managers, policy makers, researchers, and field officers. Key among these challenges is the need for ongoing validation of a dataset post-release (Pettorelli et al., 2014). Data users frequently require insights into the accuracy and uncertainty of model predictions for a specific temporal and spatial context of relevance to them, irrespective of the spatio-temporal context of the validation data collected prior to dataset release. Conformal prediction allows uncertainty to be quantified for retrospective studies, models, and datasets at designated regions of interest that lack inherent uncertainty estimates, as illustrated for the Dynamic World dataset. This enables data users to assess the accuracy and uncertainty of predictions for when and where it matters to them.

The inclusion of uncertainty information serves to mitigate a user's over-reliance on less confident or low-quality predictions. This integration empowers users to identify instances, down to individual pixels, where the predictive model provides accurate information. Alternatively, as part of a quality control system, for low-quality predictions, users can defer to datasets of higher signal quality, engage a field officer, or consult local experts when faced with uncertainty. Similar approaches have been proposed for high-risk AI-health applications (for example, Dvijotham et al., 2023). Here, the human-AI collaborative system learns the comparative precision of the predictive AI model in relation to a clinician's interpretation, and how that relationship fluctuates with the predictive AI's confidence scores. Thereafter, for a new patient, the system evaluates the optimal course of action, weighing the AI's decision against deferring to the clinician, with the overarching objective of determining the most accurate interpretation.

The identification of high uncertainty pixels may also expedite the transition towards collaborative human-AI systems, capitalizing on the synergistic benefits offered by each entity to overcome the limitations inherent in each individual system. The pixels that are flagged for human oversight may constitute the feedback that can contribute to an improvement cycle for the human-AI system, allowing continuous learning (Kamar, 2016; Wang et al., 2020). For example, in the demonstrative case studies, there is increased uncertainty at object boundaries, such as water systems or forest patches (Figure 4). Moreover, for canopy height estimation (Figure 5), the uncertainty exhibits a notable increase for pixels containing image artifacts and for high reflectance pixels in the Terene desert. In the case of the invasive tree species mapping, higher uncertainty multi-class sets that include the Dense pine class could be prioritised for subsequent labelling or field visits (Figure 6). In this way, by knowing a model's shortcomings it becomes possible to enhance system performance through augmenting data collection, rectifying inaccuracies in labelling, optimizing feature selection, or by restricting the operational domain of the system (Boulent et al., 2023; Ren et al., 2021). Such a paradigm shift toward collaborative systems is anticipated to engender enhanced trust, increased adoption rates, and overall improved operational efficiency (Dvijotham et al., 2023; Kamar, 2016; Wang et al., 2020).

In the context of the invasive species mapping case study, reference locality data were acquired from natural areas, with the deliberate exclusion of altered landscapes such as agricultural and urban environments. Consequently, increased uncertainty characterizes these anthropogenically altered regions. Conventional practice entails restricting the operational scope by masking non-natural areas. Alternatively, the Area of Applicability method takes a more nuanced approach that relies on a dissimilarity index that measures the distance of a test pixel to each instance in the models' training data, across all features (Meyer and Pebesma, 2021). However, this method has not yet gained traction in the EO community likely due to its limited availability in the R language, its lack of validity guarantees and high computational resource requirements. In contrast, the quantification of pixel-wise uncertainty provides users with the autonomy to assess if the increased uncertainty is acceptable. In the event of deemed unacceptability, users may explore alternative model or data-centric methodologies, previously mentioned, to reduce prediction uncertainty.

In instances where the consequences of omission errors bear a higher cost than including erroneous data, as exemplified in wildfire monitoring and early detection of invasive species, a judicious approach involves the retrieval of all areas potentially associated with an introduction event, subject to a user-defined confidence level (95% probability). This would be a useful way to constrain omission errors despite the accuracy of the underlying model. This property enhances the appeal of such systems for decision support frameworks relied upon by policymakers and land managers, thereby fostering the broader adoption of EO and data-driven decision-making practices.

The use of conformal prediction encourages the use of probabilistic machine learning. This may subsequently lead to positive outcomes for downstream data quality and for supporting a broader range of user needs. For example, numerous studies optimise classification systems using accuracy, precision, recall, F1-score, and more problematic metrics such as roc-auc and kappa that do not explicitly optimise model reliability since these metrics rely on discretised probability scores (Adams and Hand, 2000; Bradley, 1997; Foody, 2020; van den Goorbergh et al., 2022). In so doing, the user loses the opportunity to manually set the cost of omission and commission errors that are a function of model quality and a project's goals. This potentially leads to misalignment between the provided model outputs and the user's actual needs. Instead, binary cross-entropy (log loss) and categorical cross entropy should be preferred to avoid models that are misaligned in comparison to their objectives. Moreover, the provision of probability outputs (or at least uncalibrated pseudo-probabilities) should be encouraged (Valle et al., 2023; Venter et al., 2022), as this will allow conformal prediction to be applied independently of the data provider, and irrespective of the access to the training data and inner workings of models. Google's Dynamic World dataset is a notable example of this (Brown et al., 2022).

### 5.4. The introduced tools

To facilitate the utilization of conformal prediction methodologies by researchers and practitioners in the EO domain, we present both Python and GEE JavaScript modules as a component of an extensive mapping pipeline, demonstrated for the invasive species mapping case study. The Python modules afford dual avenues for employing conformal prediction for new studies.

First, users can use the native GEE implementations of the least ambiguous set-valued conformal classifier method for classification or the absolute residual method for regression tasks. This computational workflow was used for the dynamic world example. While we intend on supporting additional proven conformal methods natively for GEE, existing Python packages presently offer a more extensive array of approaches and may promptly accommodate new conformal prediction techniques. Consequently, we propose a second approach, wherein users can extract requisite data from GEE and seamlessly integrate it with pre-existing conformal prediction packages such as MAPIE (Taquet et al., 2022), which supports a range of conformal prediction methods. It is noteworthy that an extensive array of R, Python, and Julia packages exists to support conformal prediction; for a regularly updated compilation of resources, including textbooks, research papers, tutorials, and blogs related to conformal prediction refer to Manokhin, 2022b. We illustrated the second workflow in the context of the canopy height regression task. This workflow is preferred for regression tasks since it allows for adaptive prediction intervals through quantile regression which is not yet natively supported in GEE. The code for all three case studies is provided as annotated Jupyter notebook-based tutorials to assist the adoption and adaptation of these methods in future studies.

## 6. Future work

Contemporary trajectories in conformal prediction research revolve around the exploration of enhanced nonconformity scores that exhibit adaptability, statistical efficiency and valid coverage amidst distribution shifts, label noise (Sesia et al., 2023), missing data (Zaffran et al., 2023), nearing conditional coverage, all while maintaining computational efficiency and implementation simplicity (For example, Angelopoulos & Bates, 2023; Farinhas et al., 2023; Huang et al., 2023; Romano et al., 2019). The promise of conformal prediction and its associated validity and efficiency properties has spurred efforts among scholars to enhance established algorithms by combining them with conformal prediction. Noteworthy instances include the fusion with quantile regression (Romano et al., 2019), explainable artificial intelligence, specifically Shapley Additive exPlanations (SHAP) (Watson et al., 2023), Gaussian process regression (Papadopoulos, 2023), and Monte Carlo prediction (Bethell et al., 2023; Papadopoulos, 2023). We hope to catalyse a parallel progression within the EO domain. Candidates for integration of conformal prediction include, change point detection, anomaly detection, active learning systems and continuous monitoring algorithms such as Continual Change Detection and Classification and Continuous Degradation Detection (CCDC and CODED, respectively) (Bullock et al., 2020; Zhu et al., 2012).

## 7. Conclusion

The use of EO-derived datasets in data-driven decision-making has made a substantial contribution to the characterization, comprehension, and conservation of planet earth. Nevertheless, our examination of national to global scale datasets involved in these contributions highlights the lack of UQ accompanied by validity guarantees, and an absence of techniques capable of concurrently providing pixel-wise uncertainty information. We believe that UQ through the inclusion of conformal prediction into AI systems stands to significantly increase the role of EO data in operational monitoring systems, policy formulation, and regulatory reporting, accelerating progress towards the realisation of international planetary objectives and targets.

**CRediT authorship contribution statement**

**Geethen Singh**: Conceptualization, Methodology, Data curation, Visualization, Writing – original draft, Software. **Glenn Moncrieff**: Conceptualization, Data curation, Software, Supervision, Writing – review & editing. **Jasper Slingsby**: Writing – review & editing. **Tamara Robinson:** Funding acquisition, Project administration, Writing – review & editing, **Kerry Cawse-Nicholson**: Writing – original draft, Writing – review & editing, **Zander Venter**: Software, Writing – original draft, Writing – review & editing.

**Declaration of competing interest**

The authors declare that they have no known competing financial interests or personal relationships that could have appeared to influence the work reported in this paper.

**Declaration of AI and AI-assisted technologies in the writing process**

During the preparation of this work the authors used ChatGPT 3.5 to improve readability and language. After using this service, the authors reviewed and edited the content as needed and take full responsibility for the content of the publication.

**Acknowledgements**

This work was supported by a post-doctoral fellowship awarded to G.S. by the Centre for Invasion Biology, Stellenbosch University. KCN's contribution to this paper was carried out at the Jet Propulsion Laboratory, California Institute of Technology, under contract with the National Aeronautics and Space Administration. JS and GM were funded through National Research Foundation of South Africa (Grant Nos. 150296, 118593 and 142438). ZV was funded by the Research Council of Norway grant 160022/F40.

**Code availability**

The Python source code required for the reproduction of the three designated case studies and corresponding figures are available at the following GitHub repository: https://github.com/Geethen/GEEConformal. Within this repository, encompassed are modules and illustrative examples facilitating the application of conformal prediction methodologies to novel studies. Additionally, the Google Earth Engine (GEE) JavaScript code pertinent to conformal prediction can be accessed via the following link: https://code.earthengine.google.com/?accept_repo=users/geethensingh/conformal.

**Appendix A. Supplementary Material**

**Table A1.** A list of datasets from the Google Earth Engine (GEE) community catalogue that are not derived using machine learning and are therefore not assessed for uncertainty quantification. This list considered datasets made available up to the 2 November 2023 update.

| No. | Dataset name |
|---|---|
| 1 | GPW Version 4 adminstration units |
| 2 | geoBoundaries Global Database of Political Administrative Boundaries |
| 3 | Edge-matched Global, Subnational and operational Boundaries |
| 4 | West Africa Coastal Vulnerability Mapping |
| 5 | Social Connectedness Index (SCI) |
| 6 | Gridded Global GDP and HDI (1990-2015) |
| 7 | Harmonized Global Critical infrastructure & Index (CISI) |

| 8 | Native Land (Indigenous Land Maps) |
| --- | --- |
| 9 | Gridded Sex-Disaggregated School-Age Population (2020) |
| 10 | USA Structures |
| 11 | Geomorpho90m Geomorphometric Layers |
| 12 | Bare Earth's Surface Spectra 1980-2019 |
| 13 | Normalized Sentinel-1 Global Backscatter Model Land Surface |
| 14 | Soil nematode abundance & functional group composition |
| 15 | Global maps of habitat types |
| 16 | Global Surface water and groundwater salinity measurements (1980-2019) |
| 17 | Copernicus Digital Elevation Model (GLO-30 DEM) |
| 18 | ASTER Global Digital Elevation Model (GDEM) v3 |
| 19 | ASTER Global Water Bodies Database (ASTWBD) Version 1 |
| 20 | General Bathymetric Chart of the Oceans (GEBCO) |
| 21 | Coastal National Elevation Database (CoNED) Project -Topobathymetric digital elevation models (TBDEMs) |
| 22 | NOAA Sea-Level Rise Digital Elevation Models (DEMs) |
| 23 | ÍslandsDEM v1.0 10m |
| 24 | DEM France (Continental) 5m IGN RGE Alti |
| 25 | Soil Properties 800m |
| 26 | Polaris 30m Probabilistic Soil Properties US |
| 27 | HiHydroSoil v2.0 layers |
| 28 | Global Soil bioclimatic variables |
| 29 | Harmonized World Soil Database (HWSD) version 2.0 |
| 30 | Global Mangrove Distribution, Aboveground Biomass, and Canopy Height |
| 31 | ESA WorldCover 10 m 2020 V100 InputQuality |
| 32 | LandCoverNet Training Labels v1.0 |

| 33 | CloudSEN12 - Global dataset for semantic understanding of cloud and cloud shadow in Sentinel-2 |
|----|---|
| 34 | West Africa Land Use Land Cover |
| 35 | Mississippi River Basin Floodplain Land Use Change (1941-2000) |
| 36 | OSM Water Layer Surface Waters in OpenStreetMap |
| 37 | Global 30m Height Above the Nearest Drainage |
| 38 | Hydrography 90m Layers |
| 39 | HydroLAKES v1.0 |
| 40 | HydroATLAS v1.0 |
| 41 | HydroWaste v1.0 |
| 42 | Global River Width from Landsat (GRWL) |
| 43 | DynQual Global Surface Water Quality Dataset |
| 44 | Global coastal rivers and environmental variables |
| 45 | Global River Deltas and vulnerability |
| 46 | Streamflow reconstruction for Indian sub-continental river basins 1951–2021 |
| 47 | Global georeferenced Database of Dams (GOODD) |
| 48 | RealSAT Global Dataset of Reservoir and Lake Surface Area |
| 49 | Global Hydrologic Curve Number (GCN250) |
| 50 | Global high-resolution floodplains (GFPLAIN250m) |
| 51 | Global river networks & Corresponding Water resources zones |
| 52 | National Wetland Inventory (Surface Water and Wetlands) |
| 53 | National Hydrography Dataset (NHD) |
| 54 | Digital Earth Australia Coastlines |
| 55 | Digital Earth Africa Coastlines |
| 56 | Argo Float Data (Subset) |
| 57 | Global gridded sea surface temperature (SSTG) |
| 58 | Global Storm Surge Reconstruction (GSSR) database |

| 59 | Aqualink ocean surface and subsurface temperature subset |
| --- | --- |
| 60 | Plastic Inputs from Rivers into Oceans |
| 61 | Mismanaged Plastic Waste Dataset in Rivers |
| 62 | Global Ocean Data Analysis Project (GLODAP) v2.2022 |
| 63 | USGS VIIRS Evapotranspiration |
| 64 | USGS MODIS Evapotranspiration |
| 65 | NOAA Evaporative Stress Index (ESI) |
| 66 | Forecast Reference Crop Evapotranspiration (FRET) |
| 67 | Global Forest Carbon Fluxes (2001-2022) |
| 68 | USDA Crop Sequence Boundaries 2015-2022 |
| 69 | ESA CCI Global Forest Above Ground Biomass |
| 70 | geeSEBAL-MODIS Continental scale ET for South America |
| 71 | Global Fungi Database |
| 72 | Global Long-term Microwave Vegetation Optical Depth Climate Archive (VODCA) |
| 73 | Global Sunlit and Shaded GPP for vegetation canopies (1992-2020) |
| 74 | Aboveground carbon accumulation in global monoculture plantation forests |
| 75 | Benchmark maps Secondary Forest Brazil |
| 76 | NAFD Forest Disturbance History 1986-2010 |
| 77 | Tile Drained Croplands (30m) |
| 78 | Global crop production tillage practices |
| 79 | Global Fertilizer usage by crop & country |
| 80 | Open Aerial Map Subset |
| 81 | HySpecNet-11K Hyperspectral Benchmark dataset |
| 82 | Santa Rita Experimental Range Drone Imagery |
| 83 | USGS Historical Topo Maps |
| 84 | USGS Historical Imagery Western US |
| 85 | Global Power |
| 86 | Facebook Electrical Distribution Grid Maps |
| 87 | Harmonized Global Night Time Lights (1992-2021) |
| 88 | Global Roads Inventory Project |
| 89 | Global Highres Mining Footprints |
| 90 | Global Mining Areas and Validation Datasets |
| 91 | Global Healthsites Mapping Project |
| 92 | Global fixed broadband and mobile (cellular) network performance |
| 93 | Global Power Plant Database |
| 94 | Global offshore wind turbine dataset |
| 95 | Harmonised global datasets of wind and solar farm locations and power |
| 96 | Global Database of Cement Production Assets |
| 97 | Global Consensus Landcover |
| 98 | Global Freshwater Variables |
| 99 | Global Habitat Heterogeneity |
| 100 | Global 1-km Cloud Cover |
| 101 | Areas of global conservation value |
| 102 | CEMS Fire Danger Indices |
| 103 | Wildfire Risk to Communities (WRC) |
| 104 | Global Fire WEather Database (GFWED) |
| 105 | Global Fire Atlas (2003-2016) |
| 106 | Archival NRT FIRMS Global VIIRS and MODIS vector data |
| 107 | Monitoring Trends in Burn Severity (MTBS) 1984-2019 |
| 108 | Global large flood events (1985-2016) |
| 109 | Global Landslide Catalog (1970-2019) |
| 110 | MAXAR Open Data Events |
| 111 | Umbra SAR Open Data |
| 112 | Geocoded Disasters (GDIS) Dataset (1960–2018) |
| 113 | Global Reference Evapotranspiration Layers |
| 114 | Global Aridity Index |
| 115 | Global Wind Atlas Datasets |
| 116 | Global Solar Atlas Datasets |
| 117 | Global Extreme Heat Hazard |
| 118 | New improved Brazilian daily weather gridded data (1961–2020) |
| 119 | International Satellite Cloud Climatology Project HXG Cloud Cover |
| 120 | Current and projected climate data for North America (CMIP6 scenarios) |
| 121 | Terraclimate Individual years for +2C and +4C climate futures |
| 122 | Global MODIS-based snow cover monthly values (2000-2020) |
| 123 | MOD10A2061 Snow Cover 8-Day L3 Global 500m |
| 124 | MODIS Gap filled Long-term Land Surface Temperature Daily (2003-2020) |
| 125 | Global Daily near-surface air temperature (2003-2020) |
| 126 | Snow Data Assimilation System (SNODAS) |
| 127 | United States Drought Monitor Layers |
| 128 | North American Drought Monitor (NADM) |
| 129 | Canadian Drought Outlook |
| 130 | United States Seasonal Drought Outlook |
| 141 | Global Precipitation Measurement (GPM) |
| 142 | ANUSPLIN Gridded Climate Dataset |
| 143 | High Resolution Deterministic Prediction System (HRDPS) |
| 144 | Regional Deterministic Precipitation Analysis (RDPA) |
| 145 | Regional Deterministic Prediction System (RDPS) |
| 146 | Climate Prediction Center (CPC) Morphing Technique (MORPH) |
| 147 | Modern-Era Retrospective analysis for Research and Applications, Version 2 (MERRA2) |
| 148 | Applied Climate Information System (ACIS) NRCC NN |

| 149 | Climate Hazards Group InfraRed Precipitation with Station Data-Prelim (CHIRPS-Prelim) |
| 150 | NOAA Monthly U.S. Climate Gridded Dataset (NClimGrid) |
| 151 | High-spatial-resolution Thermal-stress Indices over South and East Asia (HiTiSAE) |
| 152 | Reference ET gridded database based on FAO Penman-Monteith for Peru (PISCOeo_pm) |
| 153 | Daylight Map Distribution map data |
| 154 | Global human modification v1.5 |

**Table A2.** A list of datasets from the Google Earth Engine (GEE) community catalogue and main catalogue (*) that are derived using machine learning but do not quantify uncertainty and are therefore not assessed for their uncertainty quantification method (Table 1). This list considered datasets made available up to the 2 November 2023 update.

| No. | Dataset name |
| --- | --- |
| 1 | High resolution settlement layer |
| 2 | Landscan |
| 3 | Relative Wealth Index (RWI) |
| 4 | Global Human Settlement Layer 2023 |
| 5 | Global ML Building Footprints |
| 6 | Global Electric Consumption revised GDP |
| 7 | Soil Organic Carbon Stocks & Trends South Africa |
| 8 | FABDEM (Forest And Buildings removed Copernicus 30m DEM) |
| 9 | Soil Grids 250m v2.0 |
| 10 | Global Soil Salinity Maps (1986-2016) |
| 11 | ESRI 10m Annual Land Use Land Cover (2017-2022) |
| 12 | GlobCover Global Land Cover |

| 13 | Finer Resolution Observation and Monitoring of Global Land Cover 10m (FROM-GLC10) |
|---|---|
| 14 | Global Impervious Surface Area (1972-2019) |
| 15 | Global urban extents from 1870 to 2100 |
| 16 | World Settlement Footprint & Evolution |
| 17 | Mapbiomas Annual land cover and use maps |
| 18 | CCI LAND COVER S2 PROTOTYPE LAND COVER 20M MAP OF AFRICA 2016 |
| 19 | South African National Land Cover (SANLC) |
| 20 | Digital Earth Australia (DEA) Landsat Land Cover 25m v1.0.0 |
| 21 | UrbanWatch 1m Land Cover & Land Use |
| 22 | Vermont High Resolution Land Cover 2016 |
| 23 | Chesapeake Bay High Resolution Land Cover Dataset (2013-2014) |
| 24 | C-CAP High-Resolution Land Cover |
| 25 | C-CAP Medium-Resolution Land Cover - Beta |
| 26 | C-CAP Wetland Potential 30m |
| 27 | Oil Palm Plantation Layers |
| 28 | Rasterized building footprint dataset for the US |
| 29 | Global River Classification (GloRiC) |
| 30 | GLOBathy (Global lakes bathymetry dataset) |
| 31 | High-Res water body dataset for tundra and boreal forests North America |
| 32 | Global Channel Belt (GCB) |
| 33 | Tensor Flow Hydra Flood Models |
| 34 | High-resolution gridded precipitation dataset for Peruvian and Ecuadorian watersheds (1981-2015) |
| 35 | Global Shoreline Dataset |
| 36 | Landfire Mosaics LF v2.2.0 |
| 37 | Vegetation dryness for western USA |

| 38 | GIMMS Normalized Difference Vegetation Index 1982-2022 |
| --- | --- |
| 39 | High-resolution annual forest land cover maps for Canada's forested ecosystems (1984-2019) |
| 40 | Canopy height forested ecosystems of Canada |
| 41 | Canada Landsat derived FAO forest identification (2019) |
| 42 | Landsat-derived forest age for Canada's forested ecosystems |
| 43 | US National Forest Type and Groups |
| 44 | Global Forest Canopy Height from GEDI & Landsat |
| 45 | Global Forest Management dataset 2015 |
| 46 | Global 30m Landsat Tree Canopy Cover v4 |
| 47 | Global tree allometry and crown architecture (Tallo) database |
| 48 | Global Leaf trait estimates for land modelling |
| 49 | NASA Harvest Layers |
| 50 | Digital Earth Africa's cropland extent map Africa 2019 |
| 51 | GFSAD Global Cropland Extent Product (GCEP) |
| 52 | GFSAD Landsat-Derived Global Rainfed and Irrigated-Cropland Product (LGRIP) |
| 53 | Global irrigation areas (2001 to 2015) |
| 54 | Global NPP-VIIRS-like nighttime light (2000-2022) |
| 55 | Global database of cement production assets and upstream suppliers |
| 56 | Global Database of Iron and Steel Production Assets |
| 57 | Biodiversity Intactness Index (BII) |
| 58 | 30m Global Annual Burned Area Maps (GABAM) |
| 59 | AgERA5 (ECMWF) dataset |
| 60 | Long-term Gap-free High-resolution Air Pollutants (LGHAP) |
| 61 | POMELO Model Population Density Maps |
| 62 | Global Intra-Urban Land Use |
| 63 | Continental-scale land cover mapping at 10 m resolution over Europe |

| 64 | *Allen coral atlas |
| 65 | *WorldPop Global Project Population Data |
| 66 | *GEOS-CF fcst tavg1hr v1 |
| 67 | *IrrMapper |
| 68 | *Dynamic world |
| 69 | *Open LandMap layers |
| 70 | *Global Human Settlement layers (GHSL) |

**Table A3.** Pre-computed qHat thresholds at various confidence levels (1-α) for Google Dynamic World based on a random 50% calibration set.

| Confidence level | qHat |
|---|---|
| 0.95 | 0.03718 |
| 0.90 | 0.06068 |
| 0.85 | 0.08787 |
| 0.80 | 0.11914 |
| 0.75 | 0.16599 |
| 0.70 | 0.22468 |